\documentclass[lettersize,journal]{IEEEtran}
\usepackage{amsmath,amsfonts}
\usepackage{algorithmic}
\usepackage{algorithm}
\usepackage{array}
\usepackage[caption=false,font=normalsize,labelfont=sf,textfont=sf]{subfig}
\usepackage{textcomp}
\usepackage{stfloats}
\usepackage{url}
\usepackage{verbatim}
\usepackage{graphicx}
\usepackage{cite}
\usepackage{tabularray}
\usepackage{placeins}
\usepackage{float}
\usepackage{hyperref}
\usepackage{xcolor}

\begin{document}

\title{Rate-Distortion Optimized Skip Coding of Region Adaptive Hierarchical Transform Coefficients for MPEG G-PCC}

\author{Zehan Wang, Yuxuan Wei, Hui Yuan~\IEEEmembership{Senior Member,~IEEE,} Wei Zhang, and Peng Li
\thanks{This work was supported in part by the National Natural Science Foundation of China under Grants 62222110, 62172259, and 62311530104, the Taishan Scholar Project of Shandong Province (tsqn202103001), the Natural Science Foundation of Shandong Province under Grant ZR2022ZD38, the State Grid Corporation Headquarters Technology Project-5500-202424166A-1-1-ZN and the OPPO Research Fund. (\textit{Corresponding author: Hui Yuan.})}
\thanks{Zehan Wang, Yuxuan Wei and Hui Yuan are with the School of Control Science and Engineering, Shandong University, Jinan 250061, China (E-mail: \protect\href{mailto:wangzehan@mail.sdu.edu.cn}{wangzehan@mail.sdu.edu.cn}; \protect\href{mailto:sduwyuxuan@mail.sdu.edu.cn}{sduwyuxuan@mail.sdu.edu.cn}; \protect\href{mailto:huiyuan@sdu.edu.cn}{huiyuan@sdu.edu.cn}).}
\thanks{Wei Zhang and Peng Li are with the School of Telecommunications Engineering, Xidian University, Xi’an, 710071, China (E-mail: \protect\href{mailto:wzhang@xidian.edu.cn}{wzhang@xidian.edu.cn};
\protect\href{mailto:penglee@mail.xidian.edu.cn}{penglee@mail.xidian.edu.cn}).}}

\markboth{Journal of \LaTeX\ Class Files,~Vol.~XX, No.~XX, July~2024}%
{Shell \MakeLowercase{\textit{et al.}}: A Sample Article Using IEEEtran.cls for IEEE Journals}

\makeatletter
\def\ps@IEEEtitlepagestyle{
  \def\@oddfoot{\mycopyrightnotice}
  \def\@evenfoot{}
}
\def\mycopyrightnotice{
  {\footnotesize
  \begin{minipage}{\textwidth}
  \centering
  Copyright~\copyright~2024 IEEE. Personal use of this material is permitted. However, permission to use this material for any other purposes \\ 
 must be obtained from the IEEE by sending an email to pubs-permissions@ieee.org.
  \end{minipage}
  }
}


\maketitle

\begin{abstract}
Three-dimensional (3D) point clouds are becoming more and more popular for representing 3D objects and scenes. Due to limited network bandwidth, efficient compression of 3D point clouds is crucial. To tackle this challenge, the Moving Picture Experts Group (MPEG) is actively developing the Geometry-based Point Cloud Compression (G-PCC) standard, incorporating innovative methods to optimize compression, such as the Region-Adaptive Hierarchical Transform (RAHT) nestled within a layer-by-layer octree-tree structure. Nevertheless, a notable problem still exists in RAHT, \textit{i.e.}, the proportion of zero residuals in the last few RAHT layers leads to unnecessary bitrate consumption. To address this problem, we propose an adaptive skip coding method for RAHT, which adaptively determines whether to encode the residuals of the last several layers or not, thereby improving the coding efficiency. In addition, we propose a rate-distortion cost calculation method associated with an adaptive Lagrange multiplier.
 Experimental results demonstrate that the proposed method achieves average Bjøntegaard rate improvements of -3.50\%, -5.56\%, and -4.18\% for the Luma, Cb, and Cr components, respectively, on dynamic point clouds, when compared with the state-of-the-art G-PCC reference software under the common test conditions recommended by MPEG. 
\end{abstract}


\section{Introduction}
\IEEEPARstart{A} three-dimensional (3D) point cloud is composed of a large number of unordered points with 3D coordinates and their associated attributes (color, reflectance, etc.). It can be used to represent 3D shape and structure of objects and scenes, and can be extensively used in immersive communication, cultural heritage preservation, and autonomous driving \cite{ref1}, etc. However, the data volume of 3D point cloud is extremely large, leading to a great challenge of efficient storage and transmission. Therefore, 3D point cloud compression was put on the agenda.

Point cloud coding methods developed significantly in recent years, including transformation-based methods \cite{ref2}, 3D-to-2D projection methods \cite{ref3}, and deep learning-based approaches \cite{ref4, ref5}. To promote the application of 3D point clouds, the Moving Picture Experts Group (MPEG) started to establish coding standards for 3D point cloud in 2017 \cite{ref6}, and proposed three test models, i.e., test model category 1 (TMC1) \cite{ref7} for static point clouds, test model category 2 (TMC2) \cite{ref8} for dynamic point clouds, and test model category 3 (TMC3) \cite{ref9} for radar point clouds. Later, as both TMC1 and TMC3 directly encode point clouds in 3D space, they are merged together and renamed as geometry-based point cloud compression (G-PCC) (the corresponding test model is named as TMC13) \cite{ref10}, while TMC2 compresses 3D point clouds by converting them into 2D geometry and attribute videos and is named as video-based point cloud compression (V-PCC) \cite{ref11}. Based on the requirement of immersive communication, to further improve the compression efficiency of solid (denser) and dynamic point clouds \cite{ref12}, MPEG proposed a new branch of G-PCC, namely, geometry-based solid content test model (GeS-TM), in 2023. This paper focuses on this new branch of G-PCC.

\begin{figure}[!t]
\centering
\includegraphics[width=\linewidth]{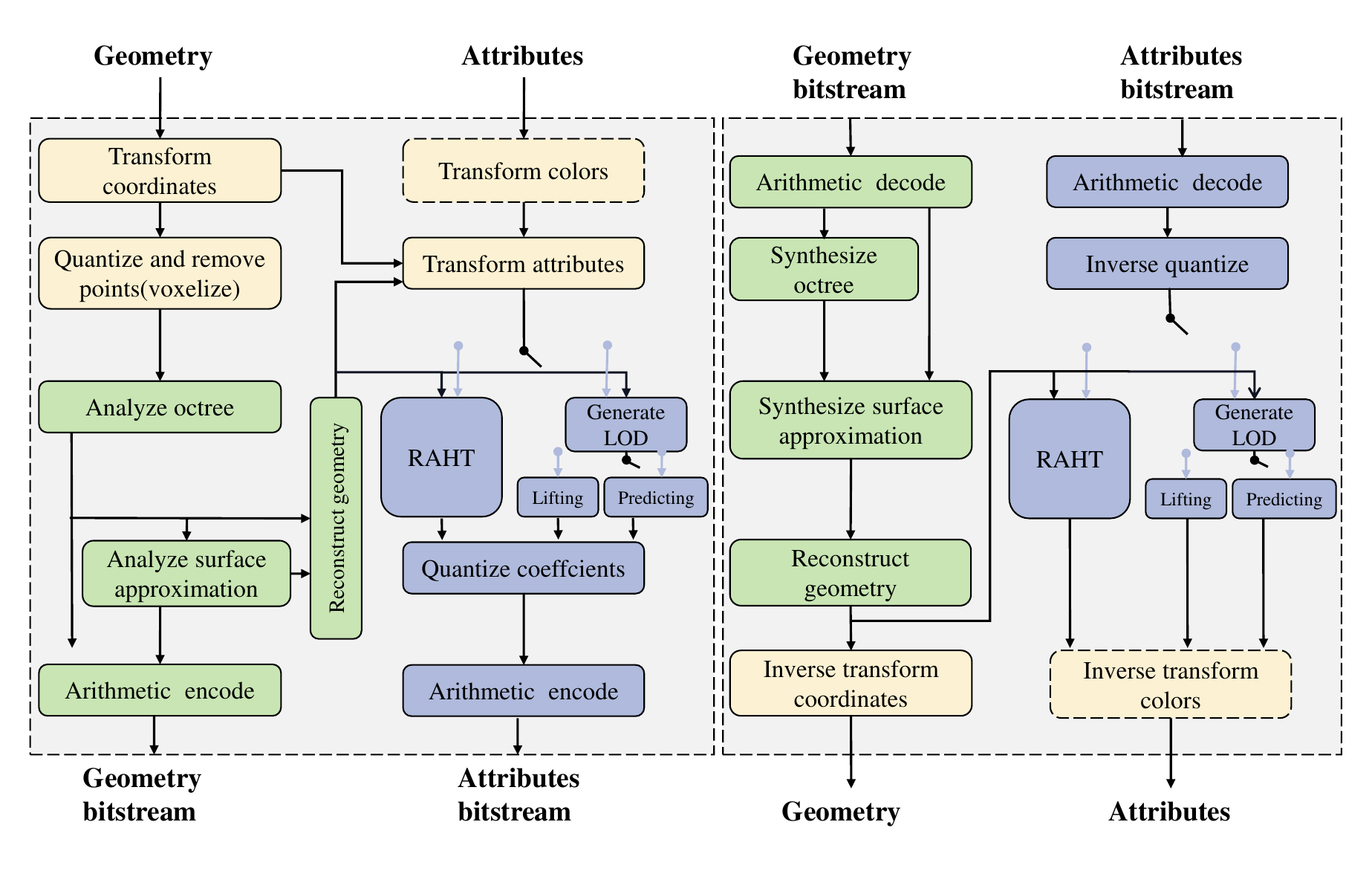}
\caption{G-PCC encoding and decoding frameworks. The yellow, green, and blue colors represent data processing, geometry processing, and attribute processing, respectively.}
\label{fig_1}
\end{figure}

The encoding and decoding frameworks of G-PCC are illustrated in Fig. \ref{fig_1}. In the encoding stage, the encoder first converts the coordinates of the input 3D point cloud to normalized coordinates that can be uniformly processed by the encoder. Subsequently, after geometry quantization and removal of redundant points, the point cloud is voxelized, and then encoded by using methods such as trisoup \cite{ref13}, octree \cite{ref14}, or predictive tree \cite{ref15}, to generate the corresponding geometry bitstream. Next, the reconstructed geometry information is used for attribute encoding in which three methods, i.e., region-adaptive hierarchical transform (RAHT) \cite{ref16}, levels of detail (LoD)-based predictive transform (PT) \cite{ref17}, and lifting transform (LT) \cite{ref18}, can be selected. Arithmetic entropy coding is then applied to the transformed coefficients to generate the corresponding attribute bitstream. In the decoding stage, the geometry bitstream is first subjected to arithmetic decoding to obtain the reconstructed geometry. The reconstructed geometry is then transformed back through inverse coordinate conversion. Simultaneously, the reconstructed geometry and attribute bitstreams are jointly fed into the decoder for arithmetic decoding and inverse quantization. Furthermore, based on the attribute encoding methods (i.e, RAHT, PT, and LT), the decoder performs corresponding inverse transformations to reconstruct the attribute information \cite{ref19, ref20}. In this paper, we focus on RAHT which is the unique transform method in GeS-TM.

RAHT is based on 3D Haar wavelet transform with an octree structure and proceeds from the root node (the upmost layer) to the leaf nodes (the bottommost layer) of the octree. The transform generates alternating-current (AC) coefficients and direct-current (DC) coefficients. The encoder encodes AC coefficients layer by layer. To reduce temporal and spatial redundancy of AC coefficients, G-PCC introduces intra-frame and inter-frame prediction for the transformed coefficients, therefore the encoder only needs to encode the residuals between the coefficients and its predicted counterpart \cite{ref21, ref22}. However, there are a large number of near-zero AC residuals in the lower layers of the octree, leading to significant redundancy in the bitstream. To further improve encoding efficiency, we propose a skip coding method for the residuals of RAHT by using Rate-Distortion (RD) Optimization (RDO) to adaptively determine whether to encode the residuals of each layer or not. Additionally, we also introduce a new rate-distortion cost calculation method associated with an adaptive Lagrange multiplier.

The rest of the paper is organized as follows. In Section II, we briefly review related work. In Section III, we describe the technical details of the proposed method. Experimental results and conclusions are given in Sections IV and V, respectively.

\section{Related Work}
\begin{figure}[!t]
\centering
\includegraphics[width=\linewidth]{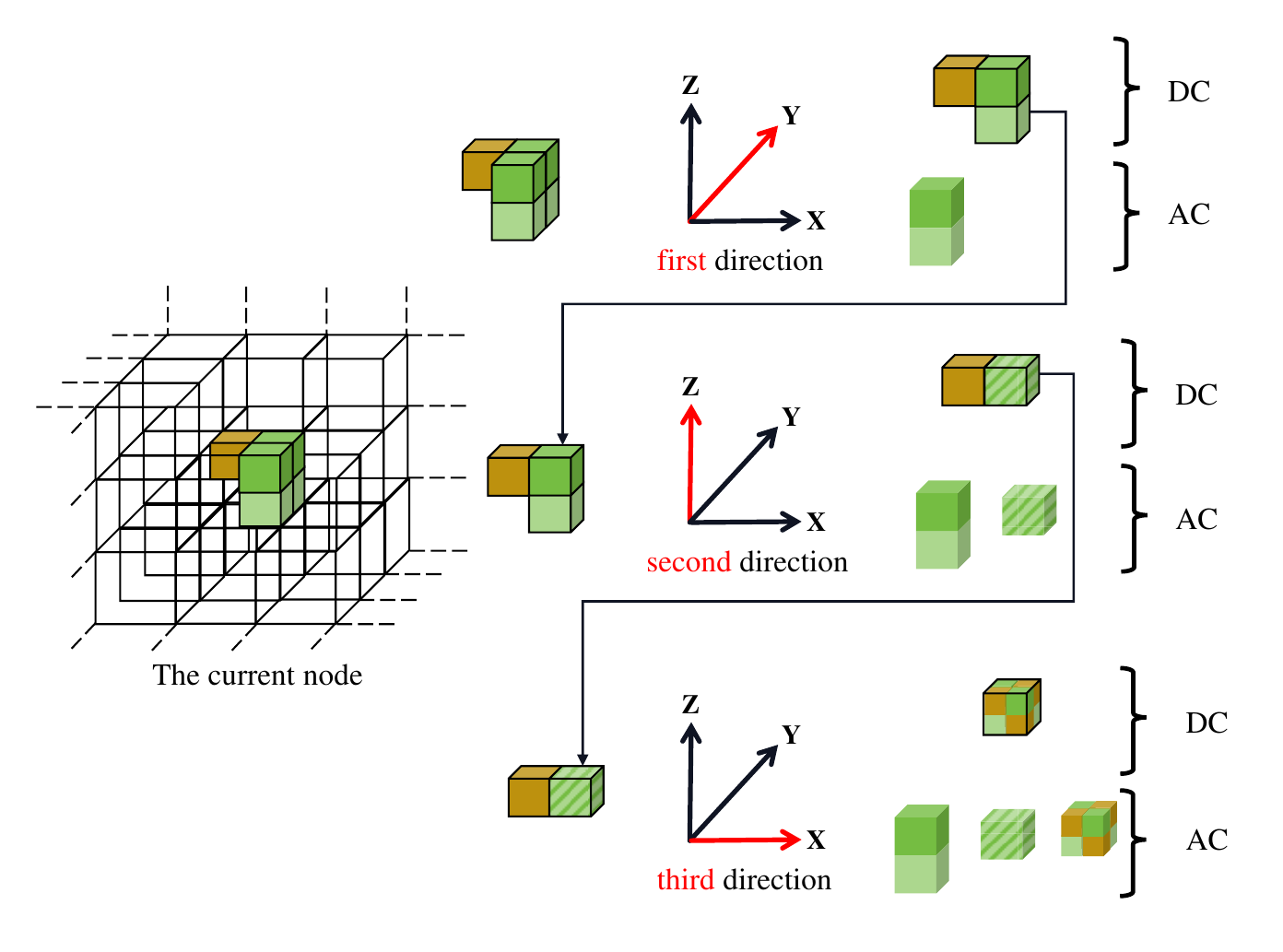}
\caption{The process of RAHT. Take the points in the current node as an example, first, a 3D Haar Wavelet transform is performed along with the Y direction to obtain 3 immediate DC coefficients and 2 AC coefficients; second, the 3 immediate DC coefficients are further transformed along with the Z direction to obtain 2 immediate DC coefficients and 1 AC coefficients; third, the 2 immediate DC coefficients are transformed along with the X direction to obtain 1 DC coefficient and 1 AC coefficient. The DC coefficient in the third stage and the AC coefficients in all the three stages are then quantized and entropy encoded.}
\label{fig_2}
\end{figure}

Research on point cloud compression has made notable progress in recent years, accompanied by ongoing enhancements and optimizations in G-PCC standard. Queiroz \textit{et al} proposed region-adaptive hierarchical transform in \cite{ref16} which was adopted by G-PCC as a transform coding method for dense point clouds. RAHT relies on a pre-partitioned octree structure, where each non-empty voxel block (namely a transform block hereafter) contains 2×2×2 sub-blocks (namely sub-blocks hereafter). Within each transform block, RAHT uses a Haar wavelet transform in the X, Y, and Z directions, respectively. The specific transformation formulas for the Haar transform of 2 sub-blocks are as follows:

\begin{equation}
\label{formula1}
\begin{bmatrix} DC \\ 
AC \end{bmatrix}=\begin{bmatrix} a & -b \\ 
a & b \end{bmatrix}\begin{bmatrix} A_{1} \\ 
A_{2} \end{bmatrix},
\end{equation}
where $ a=\sqrt{w_{1} /(w_{1}+w_{2})} $, $ b=\sqrt{w_{2} /(w_{1}+w_{2})} $, $A_{1},A_{2}$ represent the sum of attributes within the two transform sub-blocks, $w_{1},w_{2}$ represent the number of points within the two transform sub-blocks. For each transform block with 8 sub-blocks, we can obtain one DC coefficient and seven AC coefficients. The process of RAHT is illustrated in Fig. \ref{fig_2}. 

\begin{figure*}[!t]
\centering
\includegraphics[width=6.5in]{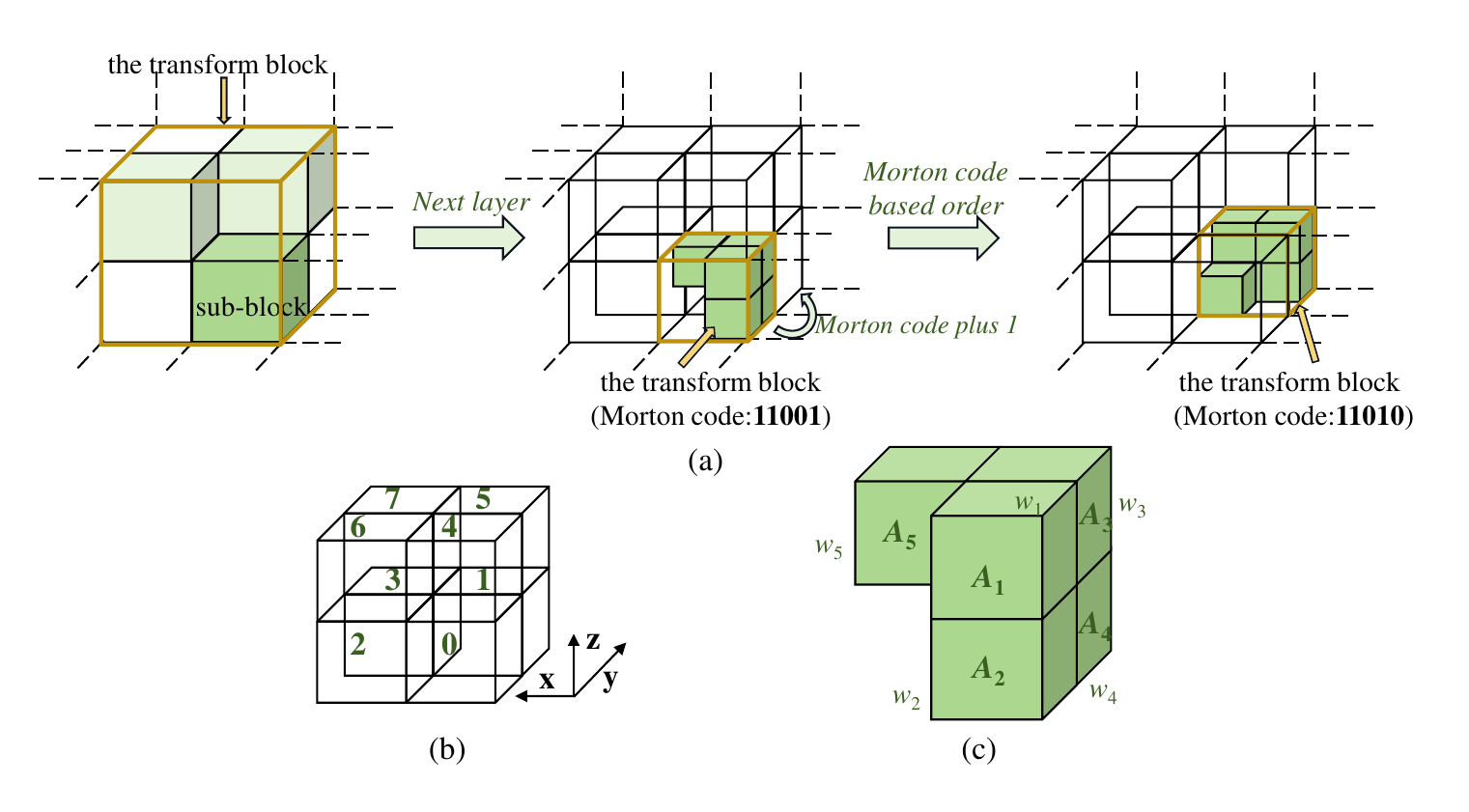}
\caption{Illustration of RAHT: (a) transform order of sub-blocks in a layer according to Morton code, (b) Morton code-based order, (c) transform results of a sub-block.}
\label{fig_3}
\end{figure*}

In the current G-PCC, the transform starts from the root node and proceeds from top to bottom. Except for the bottommost leaf nodes, the nodes in a layer can be regarded as a transform block, which contains 2×2×2 sub-blocks. Each sub-block in the current layer corresponds to the transform block of the next layer. Blocks within the same layer are transformed according to the ascending order of their block coordinates in Morton code until all 8 sub-blocks are traversed, as shown in Fig. \ref{fig_3} (a) and (b). The overall transform procedure within a transform block contains 2×2×2 sub-blocks can be simplified to a matrix operation

\begin{equation}
\label{formula2}
\begin{bmatrix}DC
 \\AC_{1} 
 \\\vdots 
 \\AC_{N-1} 
\end{bmatrix}=T(w_{1},w_{1},\dots w_{N} )\begin{bmatrix}A_{1}/\sqrt{w_{1}} 
 \\A_{2}/\sqrt{w_{2}} 
 \\\vdots 
 \\A_{N}/\sqrt{w_{N}} 
\end{bmatrix}.
\end{equation}
where $A_{1},A_{1},\dots A_{N}$ represent the sum of attributes within each sub-block, $w_{1},w_{1},\dots w_{N}$ represent the number of points within each sub-block, and $T(w_{1},w_{1},\dots w_{N})$ represents the simplified transform matrix of the transform block,  $DC,AC_{1},AC_{2},\dots A_{N-1}$ represent the transformed coefficients, as shown in Fig. \ref{fig_3}(c). During the encoding procedure, only the DC coefficient of the root node is encoded, while the DC coefficients of the other transform blocks are ignored as they can be derived from the upper layer. Therefore, only \textit{N}-1 transformed AC coefficients are encoded for the transform block.

Beyond that, recent research on RAHT has achieved significant progress. By exploring the spatial correlation of RAHT coefficients, Lasserre \textit{et al.} proposed an up-sampling-based RAHT predictive encoding method, i.e., using the attributes of the parent node and the parent neighboring node to predict the attribute of the current node \cite{ref23}. However, when there is a significant attribute variation between neighbor nodes and the current node, up-sampling-based prediction may result in significant prediction residuals. To address this problem, Zhang \textit{et al}\textit{. }proposed a threshold-based up-sampling prediction improvement in \cite{ref24}. Wang \textit{et al.} proposed to use co-plane or co-line peer neighbor attributes from sub-blocks at the same layer during the up-sampling prediction process instead of directly using the attributes of the parent neighboring nodes, to further improve prediction accuracy \cite{ref25}. To further reduce temporal redundancy of dynamic multi-frame point clouds, Xu \textit{et al.} proposed an inter-frame predictive method for RAHT transform coefficients. In this method, the corresponding node of the current node is found in the reference frame, and the attribute of this corresponding node are used to predict the attribute of the current node \cite{ref26}. When intra-frame prediction and inter-frame prediction are introduced in the RAHT algorithm, the predicted attributes are first subjected to RAHT transform to obtain the corresponding $AC_{pre}$, the original transform coefficients $AC_{org}$ are subtracted from the predicted transform coefficients $AC_{pre}$, resulting in coefficient residuals  $AC_{res}$. These residuals are then quantized and subjected to RD optimized quantization (RDOQ) \cite{ref27} followed by entropy coding. 

In lossy compression, RDO plays a pivotal role to improve the compression efficiency. It is also commonly used as an indispensable technique in V-PCC \cite{ref28, ref29, ref30} and G-PCC \cite{ref31, ref32, ref33} reference softwares. In RAHT, to better compare the performance of intra-frame prediction and inter-frame prediction, Zhang \textit{et al.} proposed an RDO-based prediction mode selection method to encode the transformed coefficients of a RAHT layer \cite{ref34}, which further improves the accuracy of RAHT coefficient prediction. To enhance the accuracy of inter-frame prediction, Ma \textit{et al}. proposed a multi-reference frame prediction scheme based on RDO selection \cite{ref35}. They adaptively select K reference frames for the current frame. These reference frames are then used to perform weighted inter-frame prediction of the RAHT coefficients for the current frame. Xu \textit{et al.} introduced a weighted average prediction between intra-frame and inter-frame predictions, and employed RDO to adaptively select the optimal prediction mode from inter-frame prediction, intra-frame prediction, and average prediction \cite{ref36}. 

\section{Proposed Method}
\subsection{Analysis of RAHT coefficients}
\begin{figure*}[ht]
\centering
\includegraphics[width=6.8in]{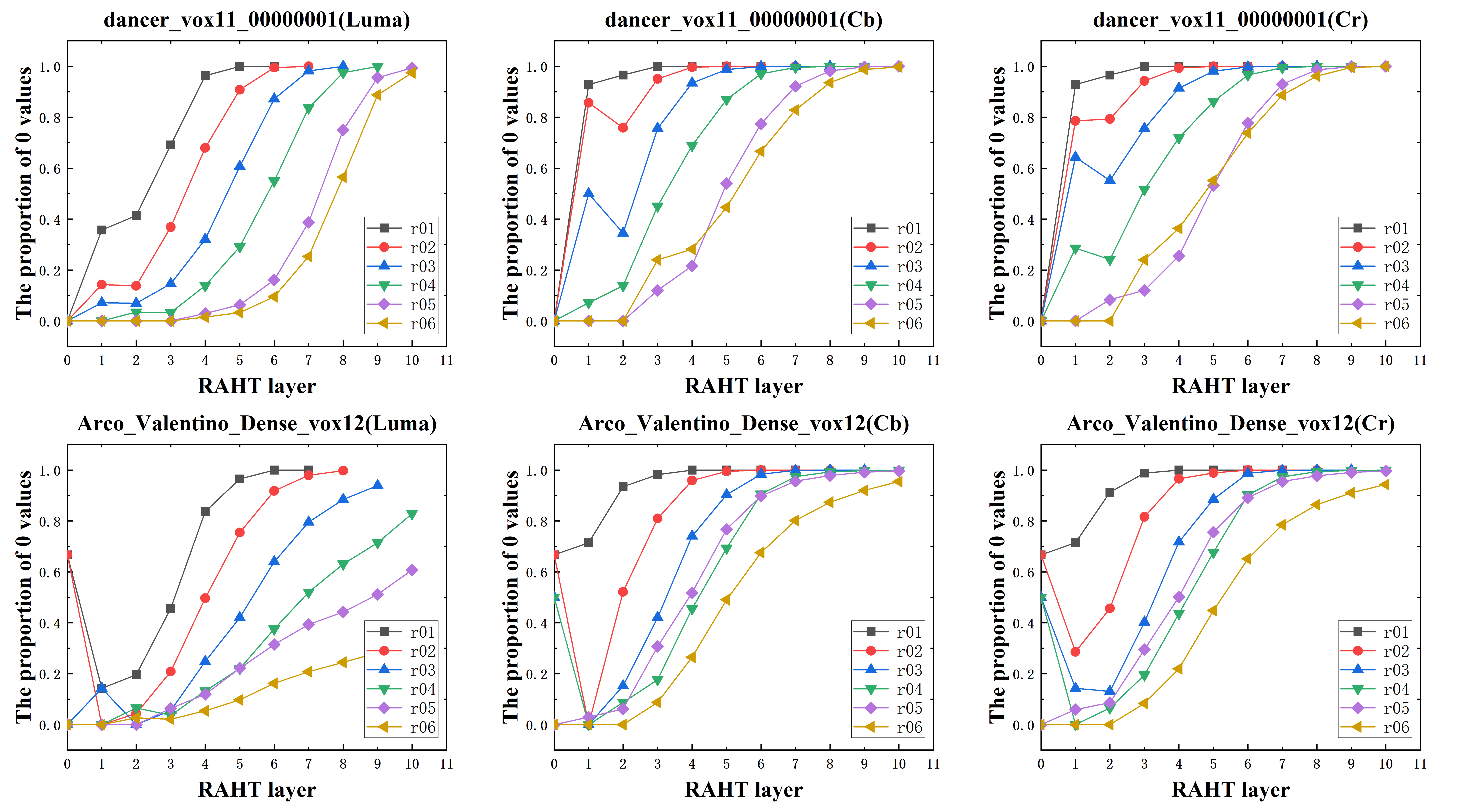}
\caption{The proportion of zero residuals to be encoded in each RAHT layer at different bitrates.}
\label{fig_4}
\end{figure*}

When conducting RAHT from top to bottom, the number of transform blocks gradually increases with the depth of layers, leading to an increasing number of AC coefficients that need to be encoded. Taking the point cloud “\textit{dancer\_vox11\_00000001}” as an example, Table~\ref{tab_1} summarizes the total number of AC coefficients that need to be encoded for each layer. Herein, layer 0 represents the root node. From the table, it is evident that, for all bitrate configurations (r01, r02, r03, r04, r05, r06, from low bitrate to high bitrate), the number of coefficients in deep layers occupy the majority of the total coefficients. Additionally, as the layers deepen, a large transform block is progressively divided into several small blocks by the octree structure, and the attributes in a large block are allocated to the small blocks, resulting in a gradual decrease of attributes within the small blocks. This, in turn, leads to a decrease of AC coefficients after RAHT of the small blocks. After the prediction and quantization, in the last few layers near the leaf nodes, the proportion of zero residuals exceeds 95\%. Fig. \ref{fig_4} illustrates the variations of zero values in each layer of the dense point cloud “\textit{dancer\_vox11\_00000001}” and the sparse point cloud “\textit{Arco\_Valentino\_Dense\_vox12}”. We can see that both dense and sparse point clouds exhibit the aforementioned trend. Particularly, for bitrates, i.e., r01 and r02, the proportion of zero values in the last two layers of Luma components and the last four layers of Chroma components exceeds 99\%. This suggests that the proportion of zero values within a layer at low bitrates is larger than that at high bitrates, and the proportion of zero values in the last few layers of Chroma components is also higher than that in Luma component. Therefore, encoding a significant amount of residuals of so many zero values in the last few layers leads to unnecessary waste of bitrate.

\begin{table}
\centering
\caption{The total number of AC coefficients to be encoded in each layer at different bitrates}
\label{tab_1}
\resizebox{\linewidth}{!}{
\begin{tabular}{c|cccccc} 
\hline
\textit{\textbf{dancer}} & \textbf{ ~r01 } & \textbf{ ~r02 } & \textbf{ ~r03 } & \textbf{ ~r04 } & \textbf{ ~r05 } & \textbf{ ~r06 }  \\ 
\hline
layer 0                                   & 2               & 2               & 2               & 2               & 2               & 2                \\
layer 1                                   & 14              & 14              & 14              & 14              & 4               & 6                \\
layer 2                                   & 29              & 29              & 29              & 29              & 12              & 10               \\
layer 3                                   & 123             & 122             & 123             & 122             & 50              & 25               \\
layer 4                                   & 511             & 547             & 549             & 549             & 204             & 135              \\
layer 5                                   & 2099            & 2177            & 2171            & 2168            & 843             & 493              \\
layer 6                                   & 8783            & 8681            & 8649            & 8610            & 3470            & 1811             \\
layer 7                                   & -               & 34575           & 34633           & 34657           & 13813           & 7538             \\
layer 8                                   & -               & -               & 135811          & 135844          & 54325           & 28866            \\
layer 9                                   & -               & -               & -               & 520522          & 211127          & 111548           \\
layer 10                                  & -               & -               & -               & -               & 788745          & 410150           \\
\hline
\end{tabular}
}
\end{table}

\subsection{Rate and distortion estimation}
To reduce the bit rate waste of the last few layers in RAHT, we propose an adaptive skip coding method for the transformed coefficients by using RDO. As the proportion of zero residuals in the last several layers is large, we first estimate the attribute bitrates of skipping all AC residuals of the last one, the last two, the last three, and the last four layers, and calculate the corresponding distortion, thus calculate the RD cost of the four cases, respectively. Then, we estimate the attribute bitrate for encoding all the AC residuals and also calculate the corresponding distortion to obtain the corresponding RD cost. Subsequently, the RD costs of the five cases are compared, and the case which gives the minimum RD cost is selected and indicated in the bit stream for the decoder. Besides, the skip coding method is independently applied to Luma, Cb, and Cr color components.

From Parseval's theorem \cite{ref37}, the distortion in the transform domain is equal to that in the attribute domain \cite{ref38}. Therefore, to save computational complexity, distortion is calculated directly in the transform domain to avoid the additional computational complexity induced by inverse RAHT transform. For the reconstructed point cloud obtained by encoding all residuals, the distortion  can be calculated as

\begin{equation}
\label{formula3}
D_{org} =\sum_{i=1}^{M_{t}} (AC_{org}^{i}- AC_{recon}^{i})^{2} ,
\end{equation}

\begin{equation}
\label{formula4}
AC_{recon}^{i}=AC_{res_q}^{i}\times Q+AC_{pre}^{i}.
\end{equation}
where $AC_{org}^{i}$ represents the \textit{i}-th AC coefficient, $AC_{recon}^{i}$ represents the reconstructed \textit{i}-th AC coefficient,  $AC_{pre}^{i}$ is the \textit{i}-th predicted AC coefficient,  $AC_{res_q}^{i}$ is the \textit{i}-th quantized AC residual, $Q$ denotes the quantization step size, and $M_{t}$ is the total number of AC coefficients. When all residuals of the last $k$, $k\in \left \{ 1,2,3,4 \right \} $, layers are skipped, all residuals of the AC coefficients in the last \textit{k} layers are ignored during encoding, they are reconstructed as zero in the decoder. Therefore, these reconstructed AC coefficients are replaced by their predicted values, and their distortion can be calculated by ${\textstyle \sum_{i=1}^{M_{k}}} (AC_{org}^{i}- AC_{pre}^{i})^{2}$. Therefore, the distortion in this case, denoted as $D_{k}$ can be represented as

\begin{equation}
\label{formula5}
D_{k} =\sum_{i=1}^{M_{k}} (AC_{org}^{i}- AC_{pre}^{i})^{2} +\sum_{i=M_{k}+1}^{M_{t}} (AC_{org}^{i}- AC_{recon}^{i})^{2}.
\end{equation}
where $M_{k}$ represents the number of AC coefficients in the last \textit{k} layers of RAHT. 

\begin{figure}[ht]
\centering
\includegraphics[width=3.5in]{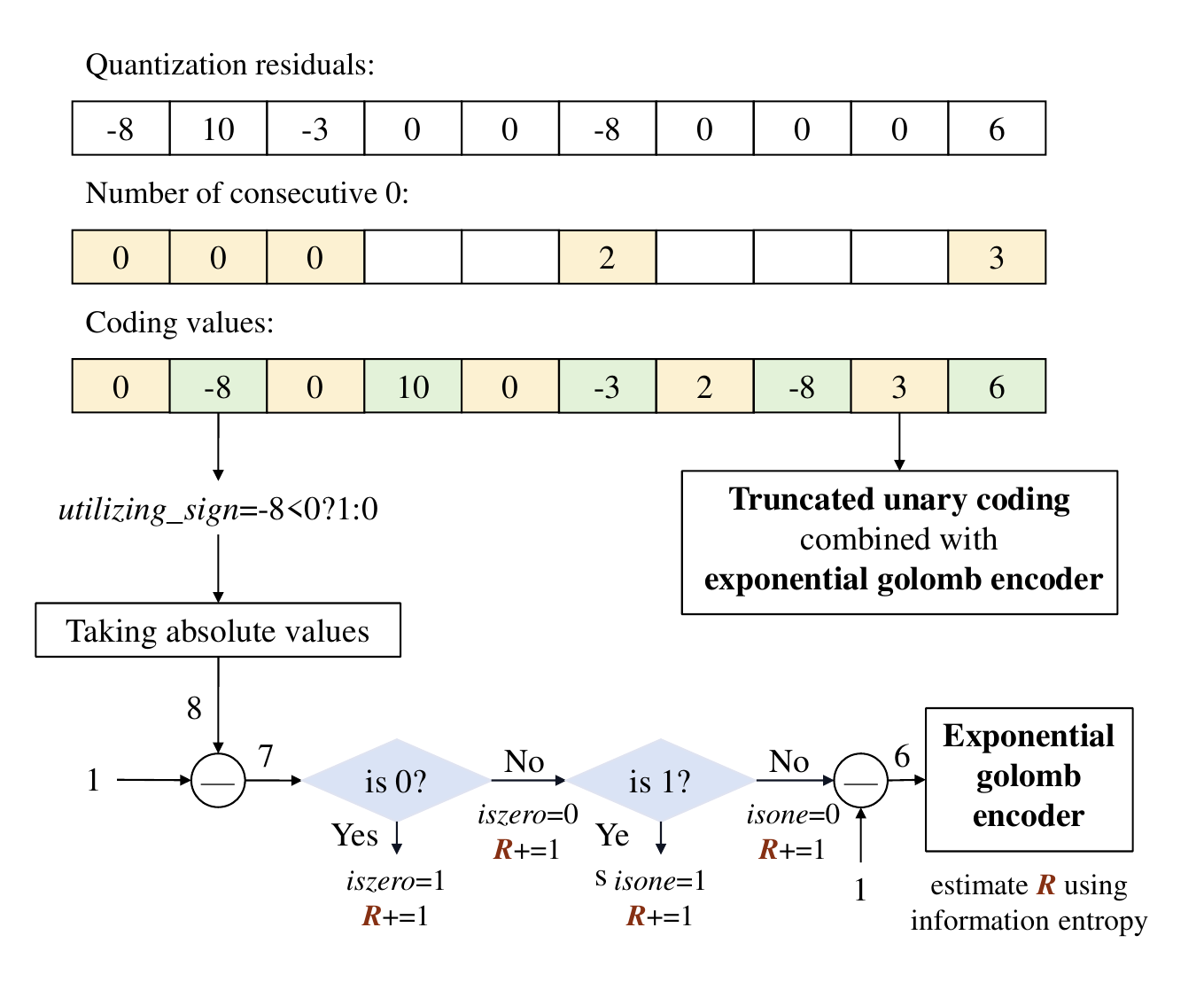}
\caption{The procedure of encoding and bitrate estimation of quantized coefficient' residuals.}
\label{fig_5}
\end{figure}
\vspace{0.1cm}

Due to the fact that most of the residuals after RAHT are zero, zero-run length coding \cite{ref39} is employed in G-PCC to encode non-zero residuals and run lengths of consecutive zeros. In practical implementation, three flags, i.e., \textit{utilizing\_sign}, \textit{iszero}, and \textit{isone} are used for non-zero residuals to distinguish negative values, 1, or 2. The remaining values are encoded using an exponential golomb encoder \cite{ref40}. For run length of consecutive zeros, truncated unary coding combined with exponential golomb encoder is used. The bitrate can be estimated by the cumulative number of bits required for encoding. For truncated unary coding, the bitrate can be estimated by the length of the codeword. For exponential golomb coding, the number of bits can be estimated by the probability of the remaining values, which is updated in real time. The procedure of encoding and bitrate estimation of quantized coefficient' residuals are illustrated in Fig. \ref{fig_5}. In the proposed method, we use $R_{org}$ to denote the attribute bits when all residuals are encoded and $R_{k}$ to represent the attribute bits (including the flags) when all AC residuals of the last $k$, $k\in \left \{ 1,2,3,4 \right \} $ layers are skipped. Therefore, the RD cost can be calculated by

\begin{equation}
\label{formula6}
RDcost=D+\lambda \cdot R,
\end{equation}
where $D$ represents the distortion ($D_{k}$ resp. $D_{org}$), $R$ represents the bits ($R_{k}$ resp. $R_{org}$), and $\lambda $ is the Lagrange multiplier which follows the basic style of H.266/VVC, i.e., 

\begin{equation}
\label{formula7}
\lambda=c\times 2^{\frac{QP-12}{3}},
\end{equation}
where $c$ is a content dependent constant and $QP$ is the quantization parameter. 

\vspace{0.1cm}
\begin{figure}[t]
\centering
\includegraphics[width=3.5in]{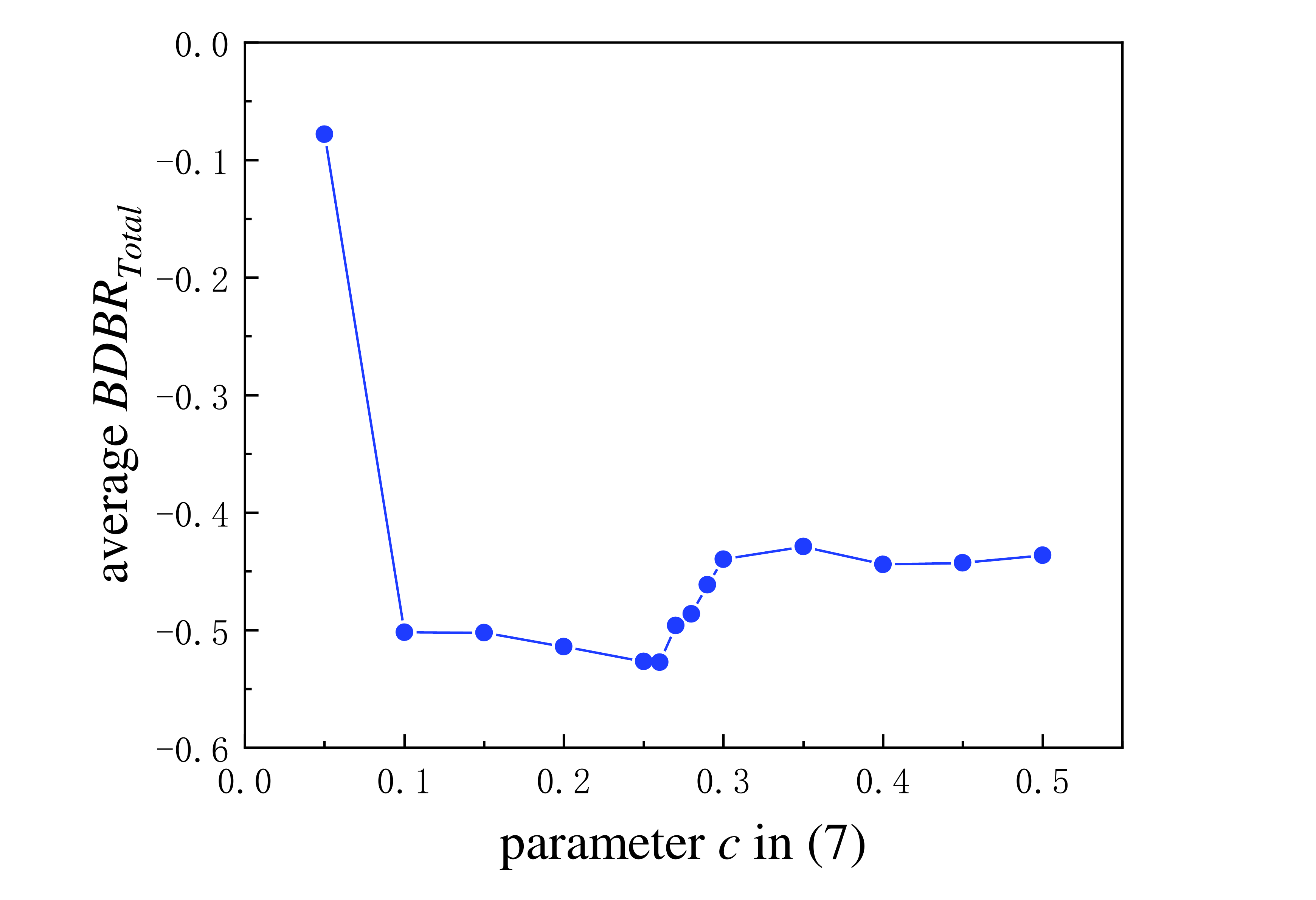}
\caption{Average $BDBR_{Total}$ variation with respective to $c$.}
\label{fig_6}
\end{figure}
\vspace{0.1cm}

\begin{table}
\centering
\caption{The $BDBR_{Total}$ with respective to  $c$ for different test point clouds}
\label{tab_2}
\resizebox{\linewidth}{!}{%
\begin{tblr}{
  cells = {c},
  vline{2} = {-}{},
  hline{1-2,16} = {-}{},
}
\textbf{constant} \textit{\textbf{c}} & \textbf{queen}     & \textbf{redandblack} & \textbf{longdress} & \textbf{basketball} & \textbf{average}  \\
0.05                & -20.69\%           & -0.80\%              & -5.87\%            & -3.92\%             & -7.82\%           \\
0.1                 & -138.95\%          & 0.53\%               & -8.76\%            & -53.53\%            & -50.18\%          \\
0.15                & \textbf{-139.17\%} & 1.31\%               & -8.37\%            & -54.61\%            & -50.21\%          \\
0.2                 & -137.20\%          & 0.12\%               & -9.42\%            & -59.13\%            & -51.41\%          \\
0.25                & -130.40\%          & -11.14\%             & -9.47\%            & -59.58\%            & -52.65\%          \\
\textbf{0.26}       & -130.44\%          & \textbf{-11.26\%}    & \textbf{-9.48\%}   & -59.83\%            & \textbf{-52.75\%} \\
0.27                & -127.86\%          & -1.60\%              & -9.41\%            & -59.56\%            & -49.61\%          \\
0.28                & -126.50\%          & -1.69\%              & -7.00\%            & -59.23\%            & -48.61\%          \\
0.29                & -126.38\%          & -0.37\%              & 1.57\%             & -59.33\%            & -46.13\%          \\
0.3                 & -126.38\%          & -0.19\%              & 10.31\%            & -59.62\%            & -43.97\%          \\
0.35                & -125.62\%          & -0.18\%              & 13.29\%            & -59.10\%            & -42.90\%          \\
0.4                 & -125.64\%          & -1.36\%              & 13.10\%            & -63.73\%            & -44.41\%          \\
0.45                & -123.05\%          & -1.20\%              & 12.50\%            & \textbf{-65.32\%}   & -44.27\%          \\
0.5                 & -120.78\%          & -0.90\%              & 12.50\%            & -65.32\%            & -43.62\%          
\end{tblr}
}
\end{table}

\begin{figure*}[ht]
\centering
\includegraphics[width=6.7in]{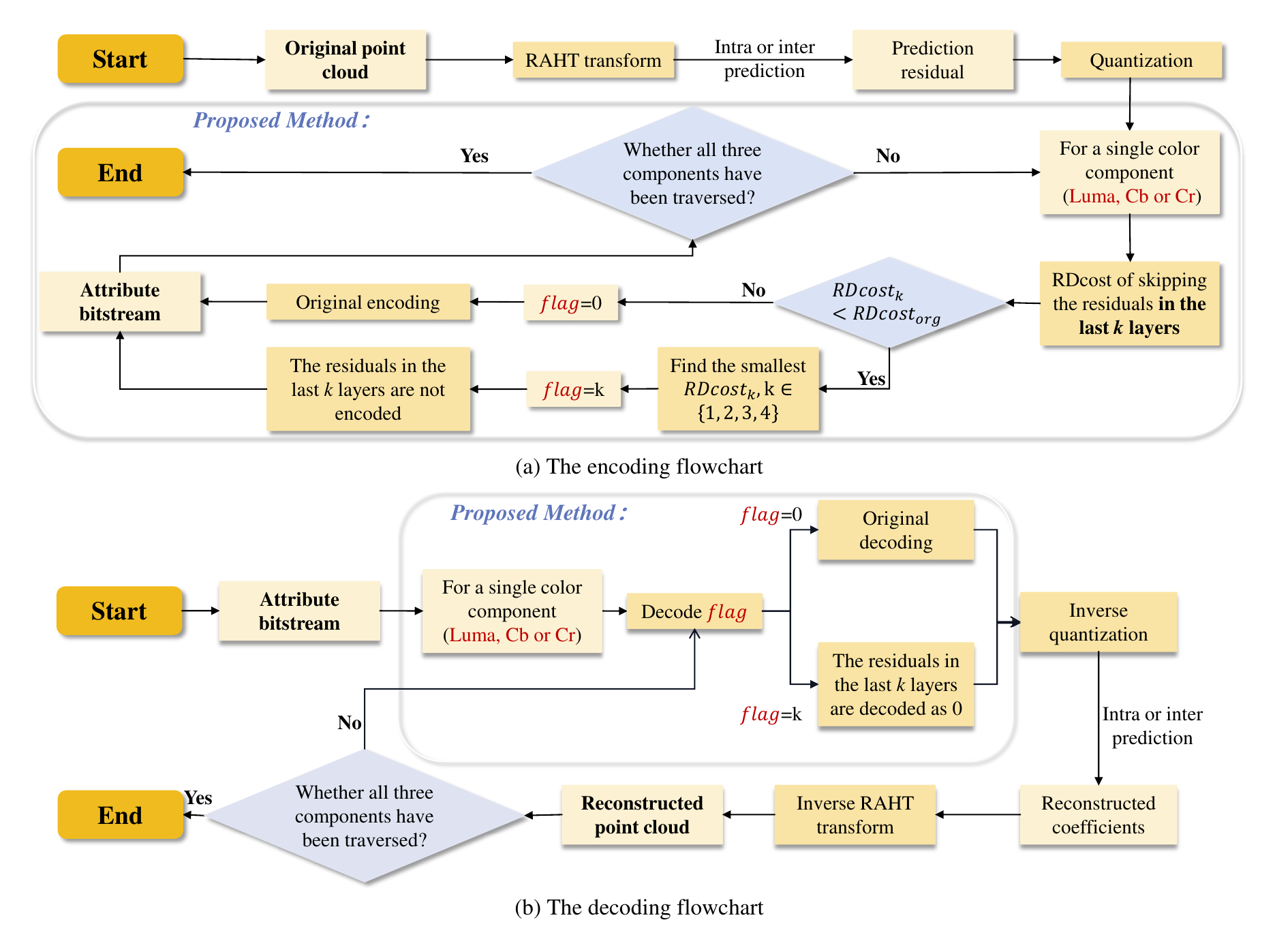}
\caption{Encoding and decoding flowcharts of the proposed method, \textit{flag }denotes \textit{flag\_luma}, \textit{flag\_cb}, and \textit{flag\_cr} for Luma, Cb, and Cr components, respectively.}
\label{fig_7}
\end{figure*}

In our study, we did extensive statistical experiments to obtain $c$. First, we implemented the proposed method onto the latest software platform GeS-TM v4.0 \cite{ref41} and encoded some typical test point clouds “\textit{queen}”, “\textit{8ivfbv2\_redandblack\_vox10}” in the dynamic point cloud category Cat2-A, “\textit{8ivfbv2\_longdress\_vox10}” in the dynamic point cloud category Cat2-B, and “\textit{basketball\_player\_vox11}” in the dynamic point cloud category Cat2-C provided by MPEG to evaluate the coding performance with respect to $c$. During the experiment, $c$ ranges from 0.05 to 0.5 with an incremental step size of 0.01. To find the optimal $c$, we use the End-to-End BD‑Attribute Rate (BDBR) \cite{ref42} to quantitatively represents the attribute bitrate savings at the same distortion when comparing the proposed method with GeS-TM v4.0. A negative BDBR indicates a positive gain compared to G-PCC, whereas a positive BDBR indicates a negative gain. We tested all the bitrates (from the lowest bitrate to the highest bitrate) by following the common test condition (CTC) \cite{ref43} of Ges-TM, and presented the average performance. By taking all the color components into account, we also define $BDBR_{Total}$ to represent the sum of BDBRs for Luma and Chroma components, 

\begin{equation}
\label{formula8}
BDBR_{Total}=a\times BDBR_{Luma}+BDBR_{Cb}+BDBR_{Cr}.
\end{equation}
where $ BDBR_{Luma}$, $ BDBR_{Cb}$ and $ BDBR_{Cr}$ represent the BDBR of Luma, Cb, and Cr components, respectively, compared to GeS-TM v4.0. The parameter $a$ equals to 7 because the importance of Luma is roughly 7 times greater than that of Chroma \cite{ref44}. The statistical results are given in Table~\ref{tab_2}, and the variation of average $BDBR_{Total}$ with respect to $c$ is shown in Fig. \ref{fig_6}. Finally, we can conclude from Table~\ref{tab_2} and Fig. \ref{fig_6} that the optimal $c$ can be set as 0.26. 

\subsection{Implementation details}
By using the optimal $c$, we can compare the $RDcost_{k}$ of skipping the residuals of the last $k$, $k\in \left \{ 1,2,3,4 \right \} $, layers with the $RDcost_{org}$ of normally encoding all the coefficients by using \eqref{formula6} and \eqref{formula7}, and pre-set a flag to indicate the best case for the decoder. If $RDcost_{org}$ is smaller than all $RDcost_{k}$, we set the flag to be $k=0$. In this case, the encoder will encode all RAHT residuals as their original values. If $RDcost_{k}$ is smaller than $RDcost_{org}$, we find the smallest $RDcost_{k}$, $k\in \left \{ 1,2,3,4 \right \} $, and set the flag as $k$ to skip the residuals of the last $k$ layers of RAHT. 

The decoder first extracts the flag from the attribute bitstream. If the flag is $k=0$, the residuals are decoded normally. If the flag is not zero, the decoder will deem all the residuals of the last $k$ layers of RAHT as zero. Subsequently, following the original G-PCC decoding process, the decoded residuals are inversely quantized to obtain $AC_{res}$. If intra-frame or inter-frame prediction was performed at the encoder side, the obtained $AC_{res}$ is added to the predicted coefficients $AC_{pre}$ to obtain the reconstructed coefficients $AC_{recon}$. Finally, inverse RAHT is applied to obtain the reconstructed attributes. The encoding and decoding framework for the proposed method can be illustrated in Fig. \ref{fig_7}. Moreover, as there are three color components, we perform the proposed method for the three color components independently and use three flags, namely \textit{flag\_luma}, \textit{flag\_cb}, and \textit{flag\_cr}, to indicate how to decode the residuals, respectively.

\section{Experimental Results and Analyses}
To verify the effectiveness of the proposed method, we implemented the proposed method in the latest test model of G-PCC reference software GeS-TMv4.0 \cite{ref41}. This test model is especially for dynamic object point clouds (categorized as Category 2, i.e., Cat2 in MPEG) which are further divided into Cat2-A, Cat2-B, and Cat2-C based on their content complexity. 

\begin{figure*} [hb]
\centering
\includegraphics[width=\linewidth]{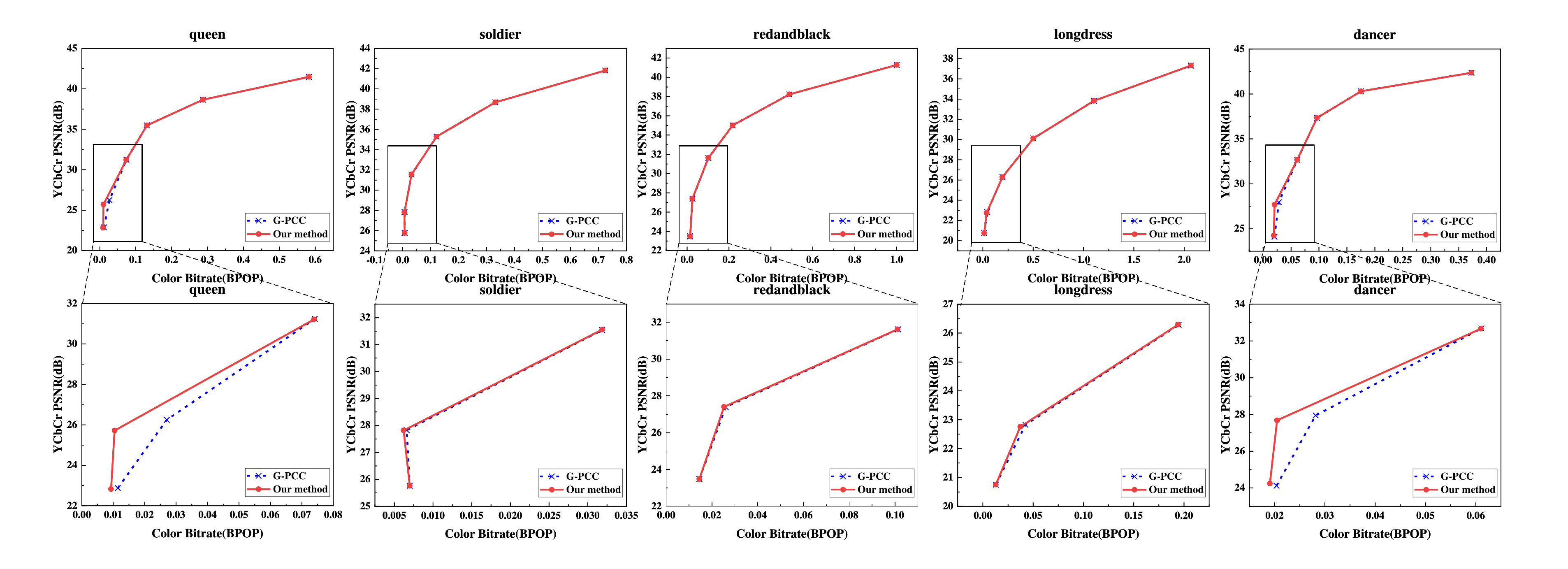}
\caption{RD curves comparison between the proposed method and G-PCC (GeS-TMv4.0).}
\label{fig_8}
\end{figure*}

The experiments were conducted according to the CTC \cite{ref43} of the testing model. Octree encoding is used for geometry, while RAHT is used for attribute encoding. We tested two conditions: lossless geometry lossy attributes (defined as C1 condition) and lossy geometry lossy attributes (defined as C2 condition), by following the CTC. When using GeS-TM for dynamic point cloud compression, inter-frame prediction was also enabled, namely “\textit{Octree-RAHT-inter}”. We also conducted the experiment when inter-frame prediction was disabled under the C1 and C2 condition, namely “\textit{Octree-RAHT-intra}” for extensive comparison and analyses. The hardware platform was an Intel I7-8700K CPU with 16GB of memory, and the software platform was the Windows 10 operating system.

\begin{table}[h]
\centering
\caption{BD-rates when comparing with GeS-TMv4.0-Octree-RAHT-inter under C1 and C2 condition}
\label{tab_3}
\resizebox{\linewidth}{!}{%
\begin{tblr}{
  cells = {c},
  cell{1}{1} = {r=2}{},
  cell{1}{2} = {c=6}{},
  cell{2}{2} = {c=3}{},
  cell{2}{5} = {c=3}{},
  vline{5} = {2-14}{},
  vline{2} = {1-14}{},
  hline{1,4,15} = {-}{},
  hline{2} = {2-7}{},
}
\textbf{Octree-RAHT-inter} & \textbf{End-to-End BD‑AttrRate (\%)} &                   &                  &                       &                   &                   \\
                           & \textbf{C1 condition}                &                   &                  & \textbf{C2 condition} &                   &                   \\
\textbf{Class}             & \textbf{Luma}                        & \textbf{Cb}       & \textbf{Cr}      & \textbf{Luma}         & \textbf{Cb}       & \textbf{Cr}       \\
loot                       & -0.23\%                              & 0.29\%            & -0.18\%          & -0.33\%               & -0.57\%           & -0.83\%           \\
redandblack                & 0.16\%                               & 0.09\%            & 0.13\%           & -0.97\%               & -2.49\%           & -0.59\%           \\
soldier                    & -0.03\%                              & -0.10\%           & -0.05\%          & -1.23\%               & -3.71\%           & -1.97\%           \\
queen                      & \textbf{-3.04\%}                     & \textbf{-10.30\%} & \textbf{-4.01\%} & \textbf{-14.35\%}     & \textbf{-20.68\%} & \textbf{-18.15\%} \\
longdress                  & 0.21\%                               & -0.17\%           & -0.39\%          & -1.04\%               & -1.71\%           & -1.89\%           \\
basketball\_player         & -0.35\%                              & -0.47\%           & -1.58\%          & -1.44\%               & 1.54\%            & 1.04\%            \\
dancer\_player             & 0.08\%                               & 0.21\%            & 1.29\%           & -5.14\%               & -11.27\%          & -6.87\%           \\
Cat2-A average             & -0.78\%                              & -2.50\%           & -1.03\%          & -4.22\%               & -6.86\%           & -5.39\%           \\
Cat2-B average             & 0.21\%                               & -0.17\%           & -0.39\%          & -1.04\%               & -1.71\%           & -1.89\%           \\
Cat2-C average             & -0.13\%                              & -0.13\%           & -0.14\%          & -3.29\%               & -4.87\%           & -2.91\%           \\
Overall average            & \textbf{-0.46\%}                     & \textbf{-1.49\%}  & \textbf{-0.68\%} & \textbf{-3.50\%}      & \textbf{-5.56\%}  & \textbf{-4.18\%}
\end{tblr}
}
\end{table}

\subsection{BD-rate Comparison under CTC}

In the field of 3D point cloud compression standardization, Bits Per Output Point (BPOP) is usually adopted to measure the average bitrate required for each point. For the same reconstruction quality, a lower BPOP indicates higher encoding efficiency. The reconstruction quality of a point cloud is usually measured by Peak Signal-to-Noise Ratio (PSNR), where a higher PSNR indicates higher reconstruction quality. In the CTC,  each point cloud is first encoded at 6 different quantization parameters, resulting in 6 bitrates (i.e., r01, r02, r03, r04, r05, and r06) and PSNRs. Then, Bjøntegaard rate (BD-rate, in percentage) can be calculated to quantitatively evaluate the RD performance \cite{ref42}. A negative BD-rate denotes a decrease in BPOP at the same reconstruction quality, indicating better encoding efficiency. 

\begin{table}[h]
\centering
\caption{BD-rates when comparing with GeS-TMv4.0-Octree-RAHT-intra under C1 and C2 condition}
\label{tab_4}
\resizebox{\linewidth}{!}{%
\begin{tblr}{
  cells = {c},
  cell{1}{1} = {r=2}{},
  cell{1}{2} = {c=6}{},
  cell{2}{2} = {c=3}{},
  cell{2}{5} = {c=3}{},
  vline{5} = {2-14}{},
  vline{2} = {1-14}{},
  hline{1,4,15} = {-}{},
  hline{2} = {2-7}{},
}
\textbf{Octree-RAHT-intra} & \textbf{End-to-End BD‑AttrRate (\%)} &             &             &                       &             &             \\
                           & \textbf{C1 condition}                &             &             & \textbf{C2 condition} &             &             \\
\textbf{Class}             & \textbf{Luma}                        & \textbf{Cb} & \textbf{Cr} & \textbf{Luma}         & \textbf{Cb} & \textbf{Cr} \\
loot                       & 0.04\%                               & -0.02\%     & -0.02\%     & -0.10\%               & -0.15\%     & -0.14\%     \\
redandblack                & 0.16\%                               & -0.04\%     & -0.04\%     & 0.12\%                & -0.42\%     & -0.27\%     \\
soldier                    & -0.01\%                              & -0.04\%     & -0.04\%     & -0.15\%               & -0.22\%     & -0.24\%     \\
queen                      & 0.81\%                               & -3.34\%     & 0.61\%      & 1.91\%                & -6.62\%     & -3.74\%     \\
longdress                  & 0.46\%                               & -0.64\%     & -0.20\%     & -0.04\%               & -1.23\%     & -0.04\%     \\
basketball\_player         & -0.05\%                              & -0.04\%     & -0.07\%     & 0.08\%                & -3.23\%     & -2.92\%     \\
dancer\_player             & -0.11\%                              & -0.21\%     & -0.25\%     & -0.76\%               & -4.97\%     & -4.11\%     \\
Cat2-A average             & 0.25\%                               & -0.86\%     & 0.13\%      & 0.45\%                & -1.85\%     & -1.10\%     \\
Cat2-B average             & 0.46\%                               & -0.64\%     & -0.20\%     & -0.04\%               & -1.23\%     & -0.04\%     \\
Cat2-C average             & -0.08\%                              & -0.12\%     & -0.16\%     & -0.34\%               & -4.10\%     & -3.52\%     \\
Overall average            & 0.18\%                               & -0.62\%     & 0.00\%      & 0.15\%                & -2.41\%     & -1.64\%       
\end{tblr}
}
\end{table}
\vspace{0.2cm}

\begin{figure*}[hb] 
\centering
\includegraphics[width=\linewidth]{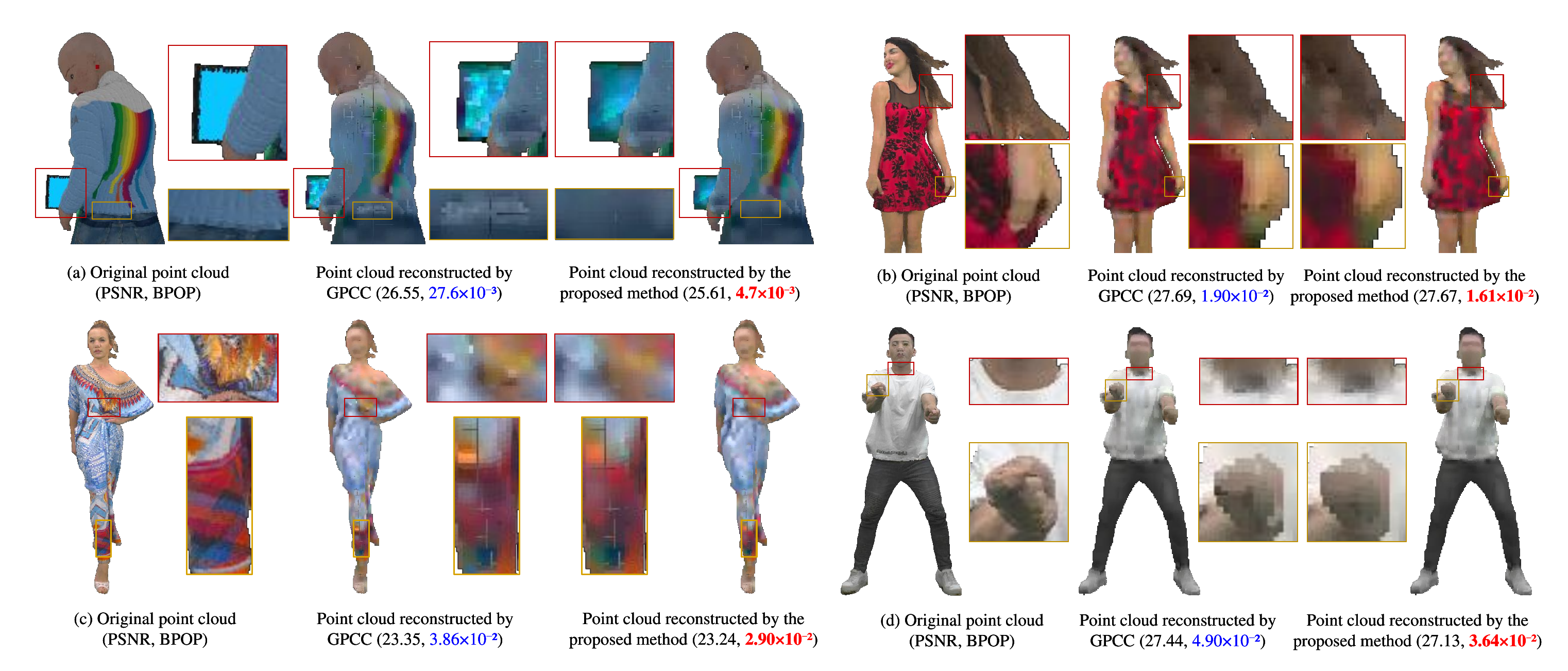}
\caption{Snapshots of dynamic point cloud sequences under C2 condition, in which the bitrate is set as r02. Left: Original point cloud. Middle: Reconstructed point cloud by G-PCC (GeS-TMv4.0). Right: Reconstructed point cloud by the proposed method. The weighted average (7:1:1) of PSNRs (dB) of Luma, Cb, and Cr are given below each sub-figure.}
\label{fig_9}
\end{figure*}

Table~\ref{tab_3} shows the performance of dynamic point clouds of the proposed method compared to GeS-TMv4.0 under the C1 and C2 conditions, where End-to-End BD-AttrRate (\%) denotes the change in attribute BPOP at the same reconstruction quality. Table~\ref{tab_4} gives the results when attribute inter-frame prediction is disabled and only intra-frame prediction is used under the C1 and C2 conditions.

From Table~\ref{tab_3}, we can see that the RD performance gain of C2 condition is larger than that of C1 condition. This is because the number of points under the lossless geometry compression of C1 condition leads to larger total coding bits, and therefore, the ratio of saved bits by the proposed method is not so significant. Besides, in C1 condition, the values of AC residuals in the last few RAHT layers are also large, leading to more reconstructed distortion and PSNR loss if they are skipped. Specifically, at C1 condition, the average BD-rates for Luma, Cb, and Cr components achieves -0.46\%, -1.49\%, and -0.68\%, respectively, while at C2 condition, the corresponding average BD-rates are -3.50\%, -5.56\%, and -4.18\%, respectively. It is noteworthy that the average BD-rates of the three color components achieve -14.35\%, -20.68\%, and -18.15\%, respectively, for “\textit{queen}”. This is because “\textit{queen}” is a computer-generated imagery (CGI) sequence, which has stronger inter-frame correlation compared to regular sequences, making the color attributes much easier to be predicted within the RAHT. Therefore, when inter-frame prediction is enabled, a larger proportion of the quantized AC coefficient' residuals are zero values. Accordingly, using the proposed method achieves for significant bitrate savings. From Table~\ref{tab_4}, we can also get the conclusion that the RD performance gain of C2 condition is larger than that of C1 condition. 

By comparing Table~\ref{tab_3} with Table~\ref{tab_4}, we can observe that the proposed method achieves better performance when inter-frame prediction is enabled. This is because, when we skipped the residuals in the last few layers, $AC_{pre}$ is directly used to represent $AC_{recon}$ on the decoder side. The accuracy of inter-prediction is typically better than intra-prediction, resulting in higher PSNR and RD performance.  

\subsection{RD Comparison}

To analyze the reconstruction quality of point clouds at different bitrates, we conducted an analysis on the dynamic point cloud “\textit{queen}”, “\textit{8ivfbv2\_soldier\_vox10}”, “\textit{8ivfbv2\_redandblack\_vox10}” in Cat2-A category, “\textit{8ivfbv2\_longdress\_vox10}” in Cat2-B category, and “\textit{dancer\_player\_vox11}” in Cat2-C category under the “\textit{Octree-RAHT-inter-C2}” condition. Fig. \ref{fig_8} depicts the RD curves of the proposed method and G-PCC, in which the weighted average (7:1:1) of PSNR for Luma, Cb, and Cr components is used. We can see that the proposed method exhibits better performance at lower bitrates. This is because lower bitrates entail a larger quantization step, leading to a substantial number of zero values in the last few layers of AC residuals in RAHT. Therefore, significant bit rate saving and minimal PSNR loss can be achieved by the proposed method. At high bit rates, as the significant distortion leads to a high RD cost, the proposed method is hard to be selected by RDO.

\subsection{Snapshots of Dynamic Sequences}

Due to the difficulty of visualizing dynamic point cloud sequences, we randomly selected certain snapshots from the dynamic point cloud sequences. By comparing the subjective quality of these snapshots, we aim to approximately evaluate the performance of our method. Fig. \ref{fig_9} shows some snapshots of the point clouds reconstructed by the proposed method and GesTMv4.0. Specifically, we selected the point cloud “\textit{queen\_frame\_1.ply}”, “\textit{redandblack\_frame\_8.ply}”, “\textit{longdress\_vox10\_frame\_6.ply}” and “\textit{dancer\_vox11\_frame\_2.ply}” for illustration, as shown in Fig. \ref{fig_9} (a), (b), (c), and (d). We can see that the proposed method can achieve similar subject quality by consuming much smaller bit rates. For instance, under the C2, r02 condition, the point cloud “\textit{queen\_frame\_1.ply}” was reconstructed by GPCC (GeS-TMv4.0) with a color BPOP of 27.6$\times 10^{3}$, whereas our method only required 4.7$\times 10^{3}$ color BPOP. We saved 82\% of the bitrate while achieving almost the same subjective visual quality. Besides, the proposed method can also eliminates block artifacts effectively and thus producing smoother visual appearance compared with GeS-TMv4.0. This is because the residuals in the last few layers are skipped and the AC prediction is directly used as the reconstruction, resulting in more smooth reconstruction quality.

\begin{table*}[ht]
\centering
\caption{Complexity Ratio comparison with GeS-TMv4.0}
\label{tab_5}
\begin{tblr}{
  cells = {c},
  cell{1}{1} = {c=2}{},
  cell{1}{3} = {c=2}{},
  cell{1}{5} = {c=2}{},
  cell{1}{7} = {c=2}{},
  cell{1}{9} = {c=2}{},
  cell{2}{1} = {c=2}{},
  cell{3}{1} = {r=4}{},
  cell{8}{1} = {r=2}{},
  cell{10}{1} = {c=2}{},
  vline{3,5,7,9} = {1-13}{},
  hline{1,3,11} = {-}{},
}
\textbf{Complexity Ratio (\%)} &                             & \textbf{Octree-RAHT-inter-C1 } &                 & \textbf{Octree-RAHT-intra-C1 } &                 & \textbf{Octree-RAHT-inter-C2 } &                 & \textbf{Octree-RAHT-intra-C2 } &                 \\
\textbf{Class}                 &                             & \textit{\textbf{C\textsubscript{enc}}}                & \textit{\textbf{C\textsubscript{dec}}} & \textit{\textbf{C\textsubscript{enc}}}                & \textit{\textbf{C\textsubscript{dec}}} & \textit{\textbf{C\textsubscript{enc}}}                & \textit{\textbf{C\textsubscript{dec}}} & \textit{\textbf{C\textsubscript{enc}}}                & \textit{\textbf{C\textsubscript{dec}}} \\
cat2-A                         & 8ivfbv2\_loot\_vox10        & 136.10\%                       & 101.38\%        & 165.43\%                       & 100.43\%        & 112.57\%                       & 100.33\%        & 119.25\%                       & 100.73\%        \\
                               & 8ivfbv2\_redandblack\_vox10 & 134.04\%                       & 100.02\%        & 165.05\%                       & 100.92\%        & 112.53\%                       & 100.81\%        & 119.08\%                       & 99.66\%         \\
                               & 8ivfbv2\_soldier\_vox10     & 140.66\%                       & 99.73\%         & 165.55\%                       & 100.57\%        & 114.62\%                       & 100.37\%        & 119.36\%                       & 100.26\%        \\
                               & queen                       & 142.51\%                       & 100.10\%        & 165.68\%                       & 101.19\%        & 113.86\%                       & 100.78\%        & 118.50\%                       & 100.94\%        \\
cat2-B                         & 8ivfbv2\_longdress\_vox10   & 133.13\%                       & 99.56\%         & 165.76\%                       & 100.47\%        & 112.20\%                       & 100.42\%        & 119.60\%                       & 100.84\%        \\
cat2-C                         & basketball\_player\_vox11   & 127.36\%                       & 99.93\%         & 165.51\%                       & 100.25\%        & 108.39\%                       & 100.28\%        & 115.02\%                       & 101.52\%        \\
                               & dancer\_player\_vox11       & 129.62\%                       & 99.57\%         & 166.33\%                       & 100.32\%        & 109.68\%                       & 101.56\%        & 115.31\%                       & 100.43\%        \\
Overall average                &                             & 134.87\%                       & 100.04\%        & 165.62\%                       & 100.59\%        & 112.34\%                       & 100.59\%        & 113.17\%                       & 100.10\%        
\end{tblr}
\end{table*}

\subsection{Complexity Comparison}
For multi-frame dynamic point clouds, the average encoding and decoding complexity \( C \) is formulated according to:
\begin{equation}
\label{formula9}
C_{enc/dec} = \frac{T_{end/dec}^{pro} }{T_{end/dec}^{anc}} \times 100\%,
\end{equation}
where $T_{end/dec}^{pro}$ and $T_{end/dec}^{anc}$ represent the average encoding/decoding time of the proposed method and anchor (i.e., G-PCC), respectively. Table~\ref{tab_5} shows the compares time complexity of the proposed method with GeS-TMv4.0 under the C1 and C2 conditions, respectively. Compared to the GeS-TMv4.0, there is a certain increase in encoding time for the proposed method, however, the decoding time remains more or less the same. We can also see that the encoding time increase of the proposed method under C1 condition is greater than that under C2 condition. This is because, during RDO, distortion and bitrate are calculated block by block. As the number of blocks is quite large in lossless geometry conditions (C1 condition), leading to higher time complexity. Additionally, the increase of encoding time under inter-frame coding configuration is less than that under intra-frame coding configuration, as inter-frame prediction results in smaller AC coefficients' residuals (i.e., more zero residuals), making faster distortion and bitrate estimation. Furthermore, when inter-frame prediction is enabled, the proposed method is used more frequently, reducing more time complexity of attribute reconstruction at the encoder side. We also observe a slight decrease in decoding complexity for certain point clouds. This is because the proposed method does not need to decode the residuals in the last $k$ layers.

\section{Conclusion}
We proposed a skip coding method for RAHT coefficients by adaptively determining whether to encode the residuals in the last $k$ ($k\in \left \{ 1,2,3,4 \right \} $) layers. Specifically, we proposed a comprehensive RD framework including bitrate calculation, distortion estimation, and skip coding condition for the coefficients' residuals to guarantee the RD performance. Additionally, we present an experimental method for determining the Lagrange multiplier. Experimental results demonstrate that the proposed method achieves significant improvements in encoding efficiency compared with the CTC (recommended by MPEG) of state-of-the-art G-PCC reference software (GeS-TMv4.0), especially at low bitrates. Specifically, the average BD-rates for the Luma, Cb, and Cr channels achieve -3.50\%, -5.56\%, and -4.18\%, respectively, under the configuration of lossy geometry and lossy attribute compression for inter-frame compression of dynamic point clouds. Moreover, the proposed method does not induce additional decoding complexity, which is friendly for practical applications. In the future, we will keep developing optimization technologies for MPEG G-PCC to further improve its coding efficiency at affordable complexity consumption.



\begin{IEEEbiography}[{\includegraphics[width=1in,height=1.25in,clip,keepaspectratio]{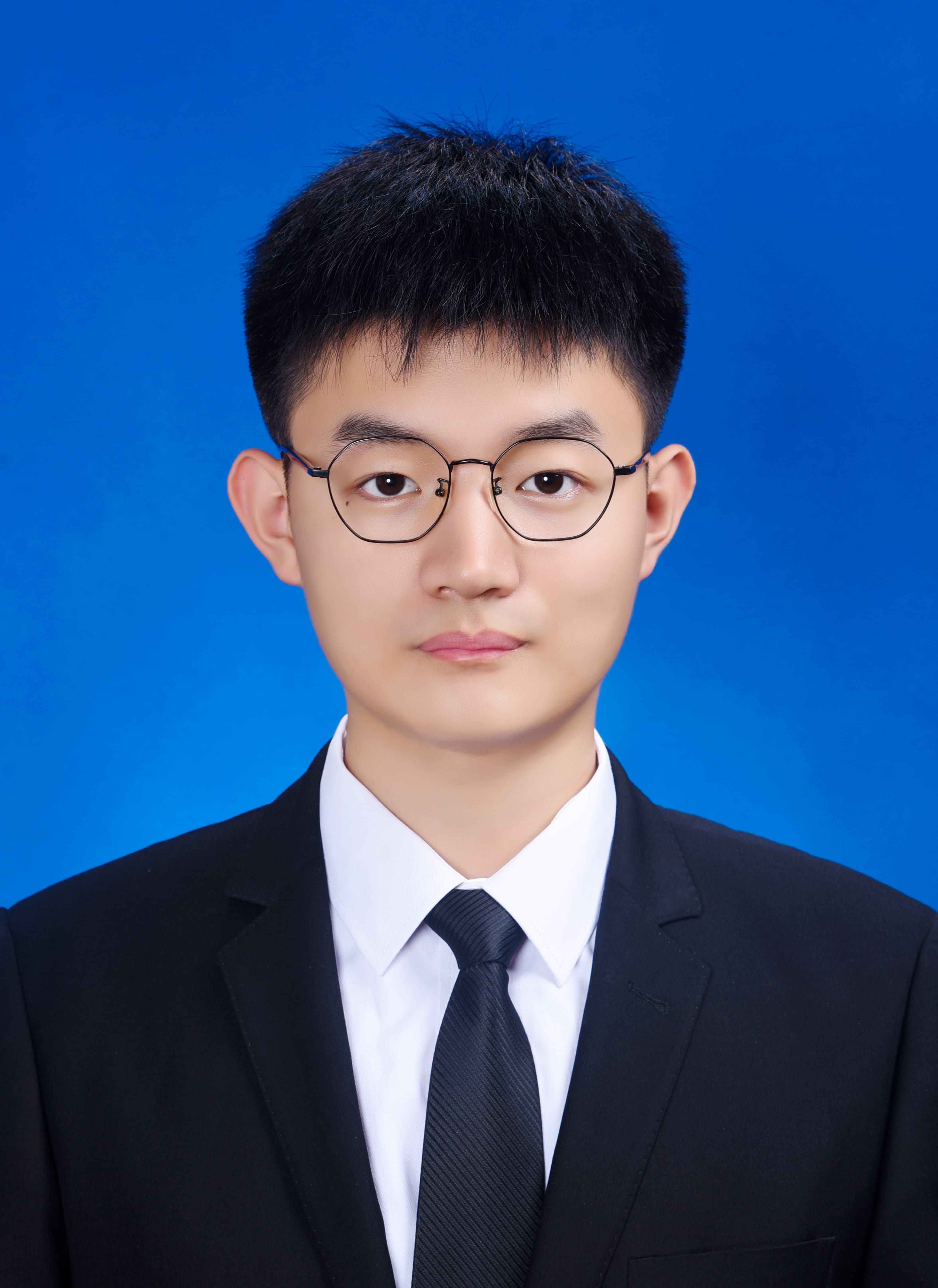}}]{Zehan Wang}
received the B.E. degree in electronic information engineering from the School of Automation and Information Engineering, Xi'an University of Technology, Xi'an, China, in 2023. He is currently pursuing the M.E. degree in the Department of Control Science and Engineering, Shandong University, Ji’nan, China. He is actively participating in the development of point cloud compression standards. His research interests include 3D point cloud compression and processing.
\end{IEEEbiography}

\begin{IEEEbiography}[{\includegraphics[width=1in,height=1.25in,clip,keepaspectratio]{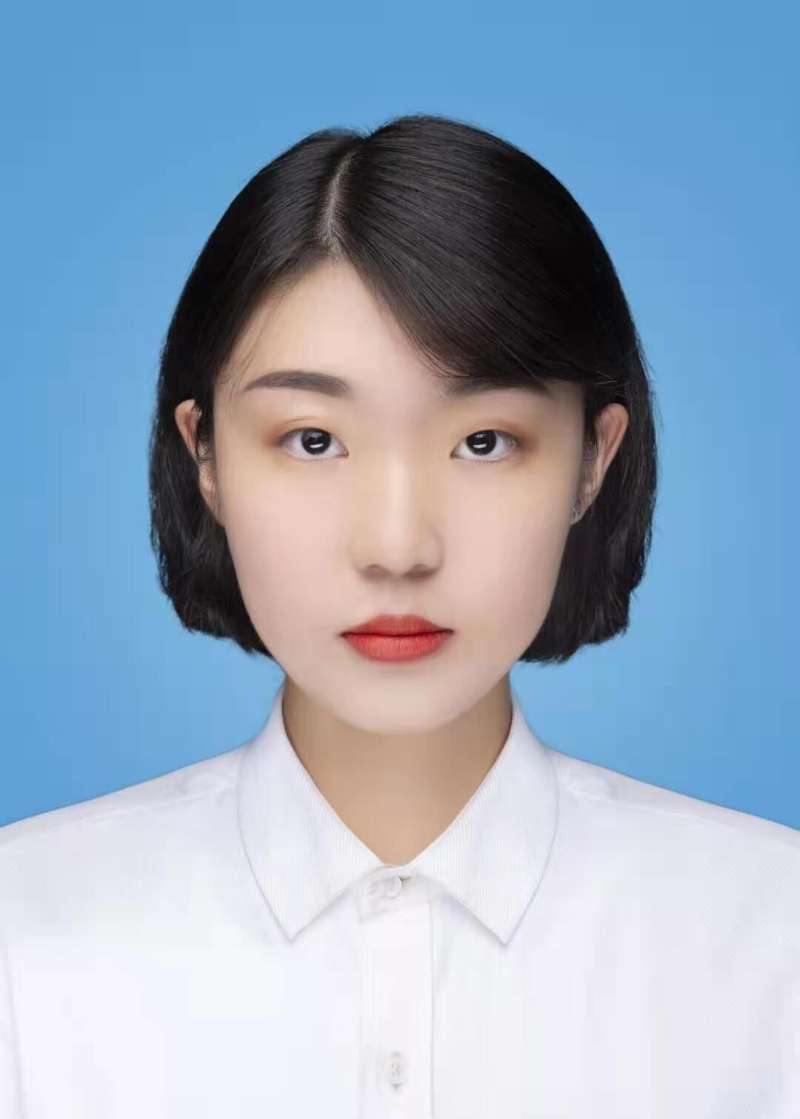}}]{Yuxuan Wei}
received the B.E. degree in automation with the Department of Control Science and Engineering from Shandong University, Ji'nan, China, in 2023. She is now pursuing the M.E. degree in Control Science and Engineering from Shandong University, Ji'nan, China. Her research interests include 3D point cloud compression and post processing.
\end{IEEEbiography}

\begin{IEEEbiography}
[{\includegraphics[width=1in,height=1.25in,clip,keepaspectratio]{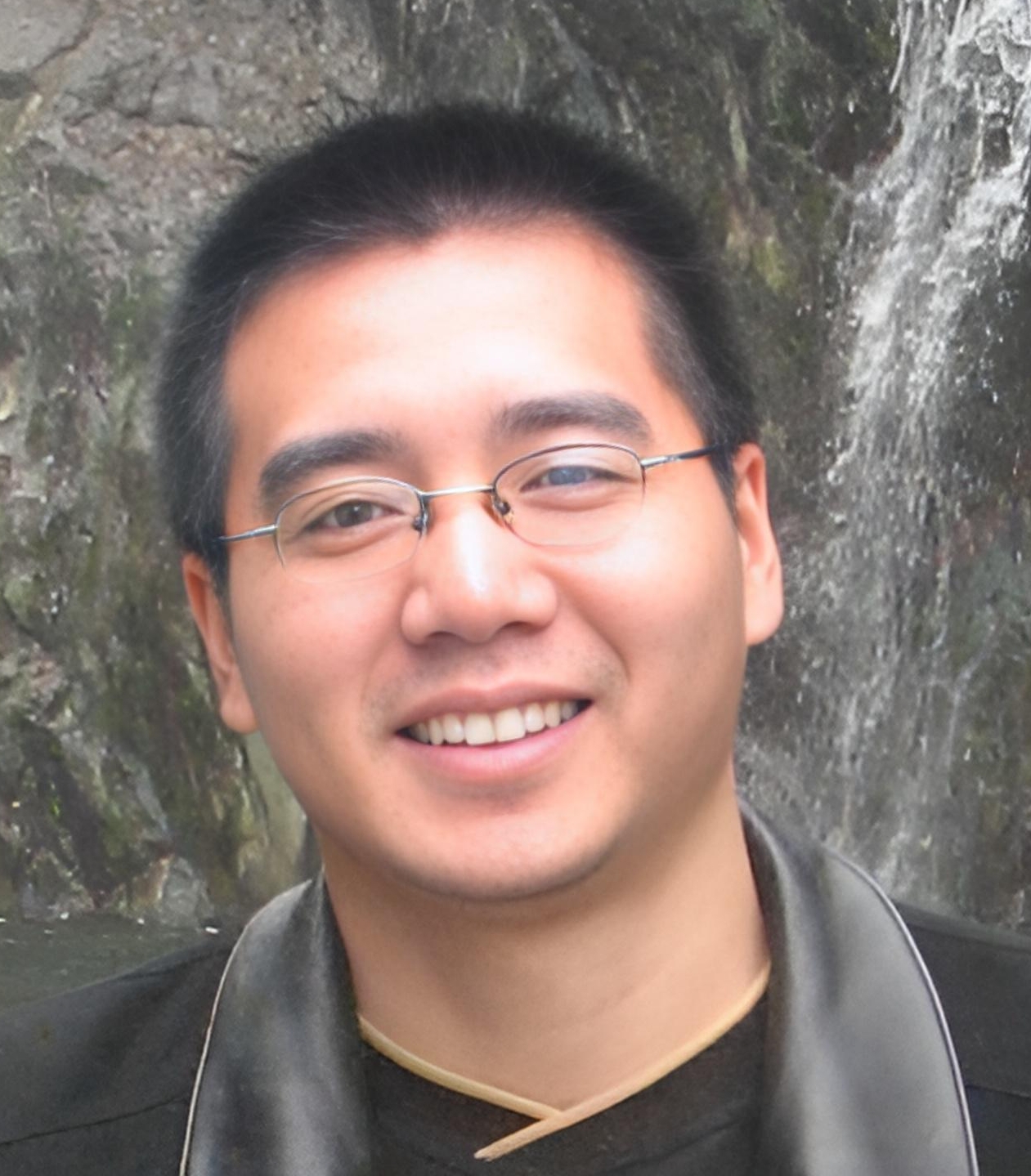}}]{Hui Yuan}
(Senior Member, IEEE) received the B.E. and Ph.D. degrees in telecommunication engineering from Xidian University, Xi’an, China, in 2006 and 2011, respectively. In April 2011, he joined Shandong University, Ji’nan, China, as a Lecturer (April 2011–December 2014), an Associate Professor (January 2015-August 2016), and a Professor (September 2016). From January 2013 to December 2014, and from November 2017 to February 2018, he worked as a Postdoctoral Fellow (Granted by the Hong Kong Scholar Project) and a Research Fellow, respectively, with the Department of Computer Science, City University of Hong Kong. From November 2020 to November 2021, he worked as a Marie Curie Fellow (Granted by the Marie Skłodowska-Curie Actions Individual Fellowship under Horizon2020 Europe) with the School of Engineering and Sustainable Development, De Montfort University, Leicester, U.K. From October 2021 to November 2021, he also worked as a visiting researcher (secondment of the Marie Skłodowska-Curie Individual Fellowships) with the Computer Vision and Graphics group, Fraunhofer Heinrich-Hertz-Institut (HHI), Germany. His current research interests include 3D visual media coding, processing, and communication.

Prof. Yuan serves as an Associate Editor for IEEE Transactions on Consumer Electronics (from 2024.06), as an Associate Editor for IET Image Processing (from 2023), an Area Chair for IEEE ICME (from 2020) and IEEE VCIP 2020, and a Senior Area Chair for PRCV 2023. He also serves as a member of IEEE CTSoc Audio/Video Systems and Signal Processing Technical Committee (AVS TC), and APSIPA Image, Video, and Multimedia Technical Committee (IVM TC).
\end{IEEEbiography}

\newpage

\begin{IEEEbiography}[{\includegraphics[width=1in,height=1.25in,clip,keepaspectratio]{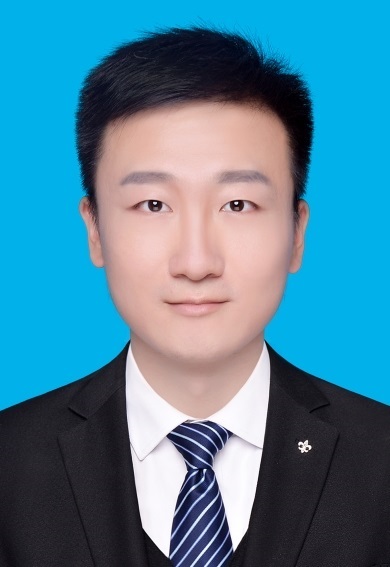}}]{Wei Zhang}
(Member, IEEE) received the B.Sc and the M.Sc degrees from Xidian University, Xi’an, China, in 2011 and 2014, respectively, and the Ph.D degree from Cardiff University, Cardiff, U.K in 2017. He is currently an Associate Professor with the School of Telecommunications Engineering, Xidian University, Xi’an, China. His research interests include visual media analysis/process and human visual perception.
\end{IEEEbiography}

\begin{IEEEbiography}[{\includegraphics[width=1in,height=1.25in,clip,keepaspectratio]{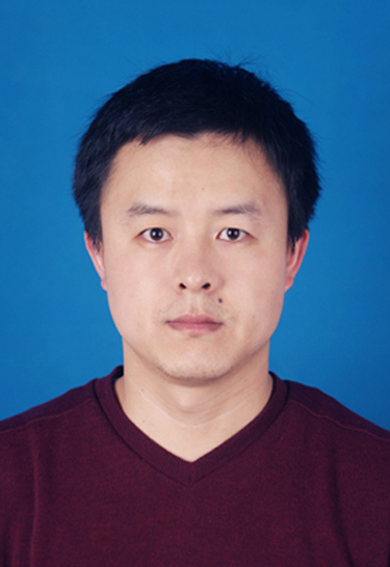}}]{Peng Li}
received the B.S. and M.S. degrees in communication engineering and the Ph.D. degree in communication and information system from Xidian University, Xi’an, China, in 2005, 2008, and 2012, respectively, where he is currently working with the State Key Laboratory on Integrated Service Networks. His research interests include image process and video streaming
\end{IEEEbiography}

\vspace{11pt}

\vfill

\end{document}